\documentclass[10pt,twocolumn,letterpaper]{article}

\usepackage{cvpr}
\usepackage{times}
\usepackage{epsfig}
\usepackage{graphicx}
\usepackage{amsmath}
\usepackage{amssymb}
\usepackage{caption}
\usepackage{subcaption}
\usepackage{multirow}
\usepackage{paralist}
\usepackage[pagebackref=true,breaklinks=true,letterpaper=true,colorlinks,bookmarks=false]{hyperref}

\def\cvprPaperID{1720} %

\cvprfinalcopy
\ifcvprfinal\fi
\begin{document}

\title{Beyond Short Snippets: Deep Networks for Video Classification}

\author{Joe Yue-Hei Ng$^1$\\
{\tt\small yhng@umiacs.umd.edu} \\
\and
Matthew Hausknecht$^2$ \\
{\tt\small mhauskn@cs.utexas.edu} \\
\and
Sudheendra Vijayanarasimhan$^3$ \\
{\tt\small svnaras@google.com} \\
\and
Oriol Vinyals$^3$ \\
{\tt\small vinyals@google.com} \\
\and
Rajat Monga$^3$ \\
{\tt\small rajatmonga@google.com} \\
\and
George Toderici$^3$ \\
{\tt\small gtoderici@google.com} \\
\and
$^1$University of Maryland, College Park\\
\and
$^2$University of Texas at Austin \\
\and
$^3$Google, Inc. \\
}

\maketitle
\begin{abstract}
Convolutional neural networks (CNNs) have been extensively applied for image
recognition problems giving state-of-the-art results on recognition, detection,
segmentation and retrieval.  In this work we propose and evaluate several deep
neural network architectures to combine image information across a video over
longer time periods than previously attempted.  We propose two methods capable of
handling full length videos. The first method explores various convolutional
temporal feature pooling architectures, examining the various design choices
which need to be made when adapting a CNN for this task. The second proposed
method explicitly models the video as an ordered sequence of frames. For this
purpose we employ a recurrent neural network that uses Long Short-Term Memory (LSTM)
cells which are connected to the output of the underlying CNN. Our best
networks exhibit significant performance improvements over previously published
results on the Sports 1 million dataset (73.1\% vs.  60.9\%) and the UCF-101
datasets with (88.6\% vs.  88.0\%) and without additional optical flow
information (82.6\% vs. 73.0\%).

\end{abstract}

\section{Introduction}
\label{sec:intro}

Convolutional Neural Networks have proven highly successful at static
image recognition problems such as the MNIST, CIFAR, and ImageNet
Large-Scale Visual Recognition Challenge \cite{krizhevsky2012imagenet,
  szegedy14going, zeiler13visualizing}. By using a hierarchy of trainable filters
and feature pooling operations, CNNs are capable of automatically learning
complex features required for visual object recognition tasks
achieving superior performance to hand-crafted features. Encouraged by these positive results
several approaches have been proposed recently to apply CNNs to video and action
classification tasks~\cite{liris2011,ji2013,karpathy2014large,simonyan2014two}.

Video analysis provides more information to the recognition task by
adding a temporal component through which motion and other information
can be additionally used. At the same time, the task is much more
computationally demanding even for processing short video clips since
each video might contain hundreds to thousands of frames, not all of
which are useful. A na\"{\i}ve approach would be to treat video frames
as still images and apply CNNs to recognize each frame and average %
the predictions at the video level. However, since each individual
video frame forms only a small part of the video's story, such an
approach would be using incomplete information and could therefore
easily confuse classes especially if there are fine-grained
distinctions or portions of the video irrelevant to the action of
interest.

\begin{figure}
\begin{center}
  \includegraphics[width=0.9\linewidth]{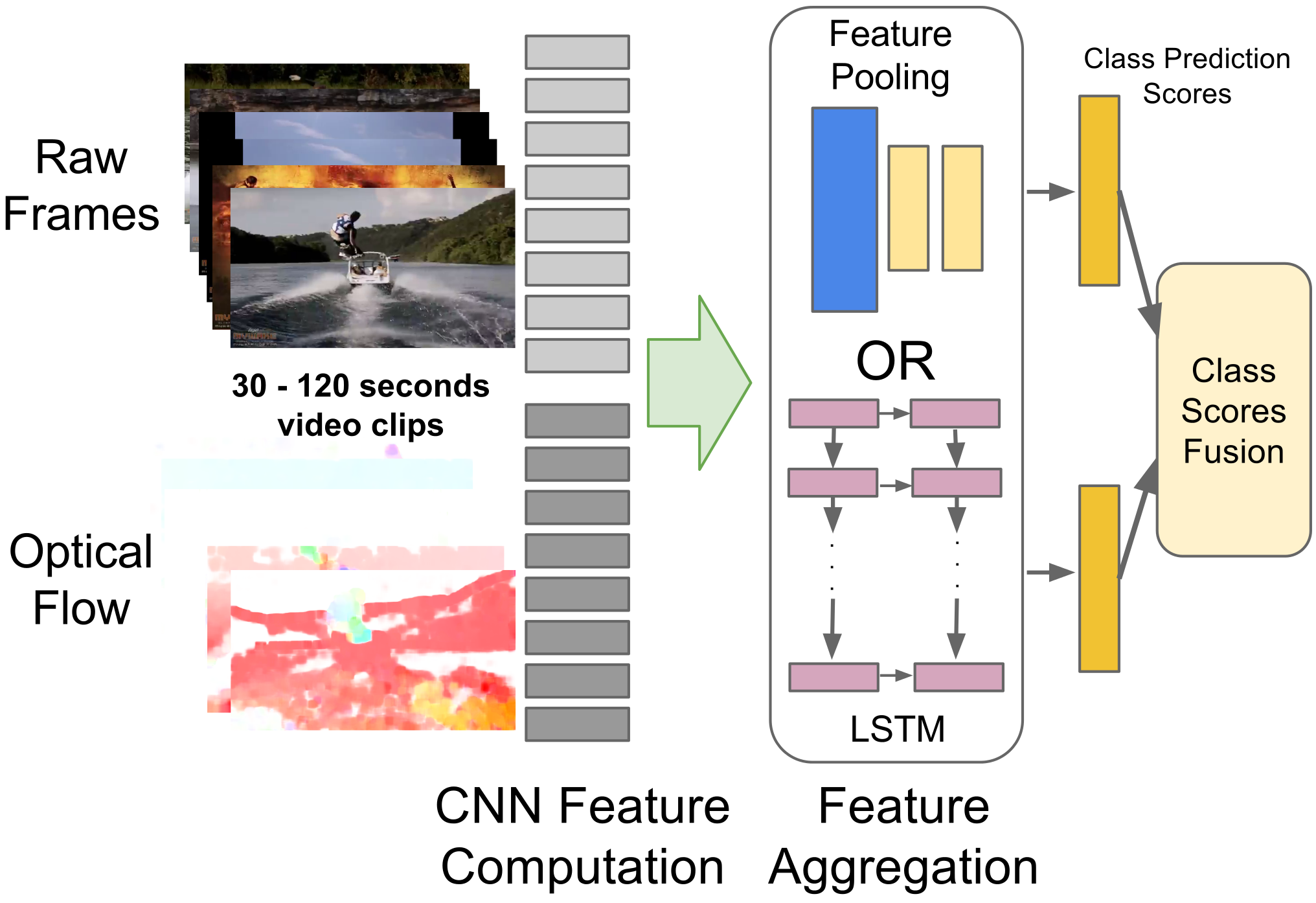}
  \caption{Overview of our approach.}
\end{center}
\end{figure}

Therefore, we hypothesize that learning a global description of the
video's temporal evolution is important for accurate video
classification. This is challenging from a modeling perspective as we
have to model variable length videos with a fixed number of
parameters.  We evaluate two approaches capable of meeting this
requirement: feature-pooling and recurrent neural networks. The
feature pooling networks independently process each frame using a CNN
and then combine frame-level information using various pooling
layers. The recurrent neural network architecture we employ is derived
from Long Short Term Memory (LSTM) \cite{hochreiter97long} units, and
uses memory cells to store, modify, and access internal state,
allowing it to discover long-range temporal relationships. Like
feature-pooling, LSTM networks operate on frame-level CNN activations,
and can learn how to integrate information over time. By sharing
parameters through time, both architectures are able to maintain a
constant number of parameters while capturing a global description of
the video's temporal evolution.

Since we are addressing the problem of video classification, it is natural to
attempt to take advantage of motion information in order to
have a better performing network. Previous work~\cite{karpathy2014large} has
attempted to address this issue by using frame stacks as input. However, this
type of approach is computationally intensive since it involves thousands of 3D
convolutional filters applied over the input volumes.  The performance grained
by applying such a method is below 2\% on the Sports-1M
benchmarks~\cite{karpathy2014large}. As a result, in this work, we avoid
implicit motion feature computation.

In order to learn a global description of the video while maintaining a low
computational footprint, we propose processing only one frame per second. At
this frame rate, implicit motion
information is lost. To compensate, following ~\cite{simonyan2014two} we
incorporate explicit motion information in the form of optical flow images
computed over adjacent frames. Thus optical flow allows us to retain the
benefits of motion information (typically achieved through high-fps sampling)
while still capturing global video information. Our
contributions can be summarized as follows:

\begin{compactenum}
\item We propose CNN architectures for obtaining global video-level
  descriptors and demonstrate that using increasing numbers of frames
  significantly improves classification performance.

\item By sharing parameters through time, the number of parameters
  remains constant as a function of video length in both the feature
  pooling and LSTM architectures.

\item We confirm that optical flow images can greatly benefit video
  classification and present results showing that even if the optical
  flow images themselves are very noisy (as is the case with the
  Sports-1M dataset), they can still provide a benefit when coupled
  with LSTMs.

\end{compactenum}

Leveraging these three principles, we achieve state-of-the-art
performance on two different video classification tasks: Sports-1M
(Section \ref{subsec:sports-1m}) and UCF-101 (Section \ref{subsec:ucf-101}).

\vspace{-.8em}
\section{Related Work}
\vspace{-1em}
\label{sec:related}

Traditional video recognition research has been extremely successful at obtaining
global video descriptors that encode both appearance and motion information in order to
 provide state-of-art results on a large number of video datasets. These approaches are able
to aggregate local appearance and motion information using hand-crafted features
such as Histogram of Oriented Gradients (HOG), Histogram of Optical
Flow (HOF), Motion Boundary Histogram (MBH) around spatio-temporal
interest points \cite{laptev-actions}, in a dense grid
\cite{Wang09evaluationof} or around dense point
trajectories~\cite{jain13,kuehne11,wang11,wang13} obtained through
optical flow based tracking. These features are then encoded in order
to produce a global video-level descriptor through bag of words
(BoW)~\cite{laptev-actions} or Fisher vector based
encodings~\cite{wang13}.

However, no previous attempts at CNN-based video recognition use both
motion information and a global description of the video: Several
approaches \cite{liris2011,ji2013,karpathy2014large} employ
3D-convolution over short video clips - typically just a few seconds -
to learn motion features from raw frames implicitly
and then aggregate predictions at the video level.
Karpathy \etal\cite{karpathy2014large} demonstrate that their network
is just marginally better than single frame baseline, which indicates
learning motion features is difficult.
In view of this, Simonyan
\etal \cite{simonyan2014two} directly incorporate motion information
from optical flow, but only sample up to 10 consecutive frames at
inference time. The disadvantage of such local approaches is that each
frame/clip may contain only a small part of the full video's
information, resulting in a network that performs no better than the
na\"{\i}ve approach of classifying individual frames.

Instead of trying to learn spatio-temporal features over small time
periods, we consider several different ways to aggregate strong CNN
image features over long periods of a video (tens of seconds)
including feature pooling and recurrent neural networks. Standard
recurrent networks have trouble learning over long sequences due to
the problem of vanishing and exploding gradients
\cite{bengio94learning}.
In contrast, the Long Short Term Memory
(LSTM) \cite{hochreiter97long} uses memory cells to store, modify, and
access internal state, allowing it to better discover long-range
temporal relationships. For this reason, LSTMs yield state-of-the-art
results in handwriting recognition \cite{graves09novel,
  graves08offline}, speech recognition \cite{graves13speech,
  graves14towards}, phoneme detection \cite{fernandez08phoneme},
emotion detection \cite{wollmer13lstm}, segmentation of meetings and
events \cite{reiter06combined}, and evaluating programs
\cite{zaremba14learning}. While LSTMs have been applied to action
classification in~\cite{Baccouche2010}, the model is learned on top of
SIFT features and a BoW representation. In addition, our proposed
models allow joint fine tuning of convolutional and recurrent parts of
the network, which is not possible to do when using hand-crafted
features, as proposed in prior work. Baccouche \etal
\cite{Baccouche2010} learns globally using Long Short-Term Memory
(LSTM) networks on the ouput of 3D-convolution applied to 9-frame
videos clips, but incorporates no explicit motion information.

\section{Approach}
\vspace{-1em}
\label{sec:approach}

\label{subsec:cnns}
Two CNN architectures are used to process individual video frames:
AlexNet and GoogLeNet. AlexNet, is a Krizhevsky-style CNN
\cite{krizhevsky2012imagenet} which takes a 220 $\times$ 220 sized
frame as input. This frame is then processed by square convolutional layers
of size 11, 9, and 5 each followed by max-pooling and local
contrast normalization. Finally, outputs are fed to two fully-connected layers
each with 4096 rectified linear units (ReLU).
 Dropout is applied to each fully-connected layer with a ratio of
 0.6 (keeping and scaling 40\% of the original outputs).

GoogLeNet \cite{szegedy14going}, uses a network-in-network approach,
stacking Inception modules to form a network 22 layers deep that is
substantially different from previous CNNs
\cite{krizhevsky2012imagenet, zeiler13visualizing}. Like AlexNet,
GoogLeNet takes a single image of size 220 $\times$ 220 as input. This
image is then passed through multiple Inception modules, each of which
applies, in parallel, 1$\times$1, 3$\times$3, 5$\times$5 convolution,
and max-pooling operations and concatenates the resulting
filters. Finally, the activations are average-pooled and output as a
1000-dimensional vector.

In the following sections, we investigate two classes of CNN architectures
capable of aggregating video-level information. In the first section, we
investigate various feature pooling architectures that are agnostic to temporal
order and in the following section we investigate LSTM networks which are capable of
learning from temporally ordered sequences. In order to make learning computationally feasible,
in all methods CNN share parameters across frames.

\vspace{-.6em}
\subsection{Feature Pooling Architectures}
\vspace{-.6em}
\label{subsec:maxpoolarch}
Temporal feature pooling has been extensively used for video
classification~\cite{laptev-actions,Wang09evaluationof,jain13}, and has been usually
applied to bag-of-words representations. Typically, image-based or motion
features are computed at every frame, quantized, then pooled across time.  The resulting
vector can be used for making video-level predictions. We follow a
similar line of reasoning, except that due to the fact that we work with neural
networks, the pooling operation can be incorporated directly as a layer. This allows
us to experiment with the location of the temporal pooling layer with respect
to the network architecture.

We analyze several variations depending on the specific pooling method and the
particular layer whose features are aggregated. The pooling operation need not
be limited to max-pooling. We considered using both average pooling, and max-pooling
which have several desirable properties as shown in~\cite{boureau10}. In
addition, we attempted to employ a fully connected layer as a ``pooling
layer''. However, we found that both average pooling and a fully
connected layer for pooling failed to learn effectively due to the large number
of gradients that they generate. Max-pooling generates much sparser updates,
and as a result tends to yield networks that learn faster, since the
gradient update is generated by a sparse set of features from each frame. Therefore, in the rest of the paper we use max-pooling as
the main feature aggregation technique.

Unlike traditional bag of words approaches, gradients coming from the top layers help learn useful features from image pixels, while allowing
the network to choose which of the input frames are affected by these updates. When used with max-pooling, this is reminiscent of multiple instance learning,
where the learner knows that at least one of the inputs is relevant to the target class.

We experimented with several variations of the basic max-pooling architecture as shown in Figure \ref{fig:maxpool}:

\begin{figure}[ht]
\begin{center}
    \begin{subfigure}[ht]{0.48\linewidth}
        \includegraphics[width=0.97\linewidth]{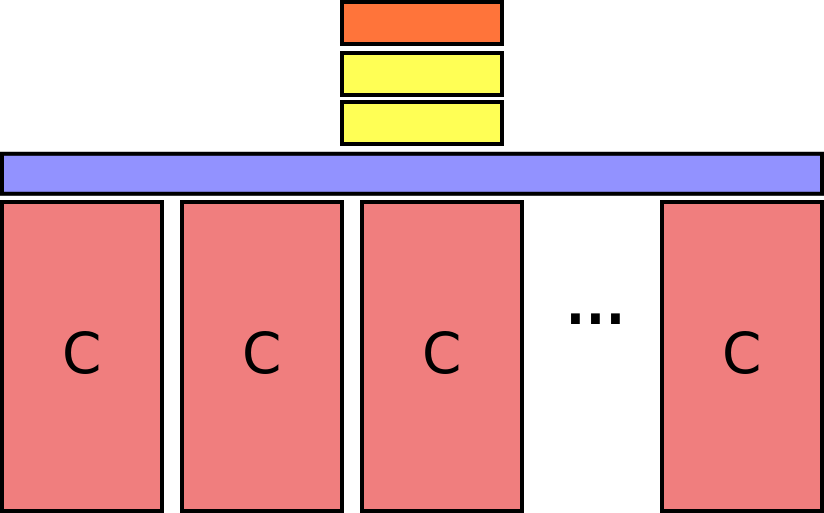}
        \caption{Conv Pooling}
    \end{subfigure}
    \begin{subfigure}[ht]{0.48\linewidth}
        \includegraphics[width=0.97\linewidth]{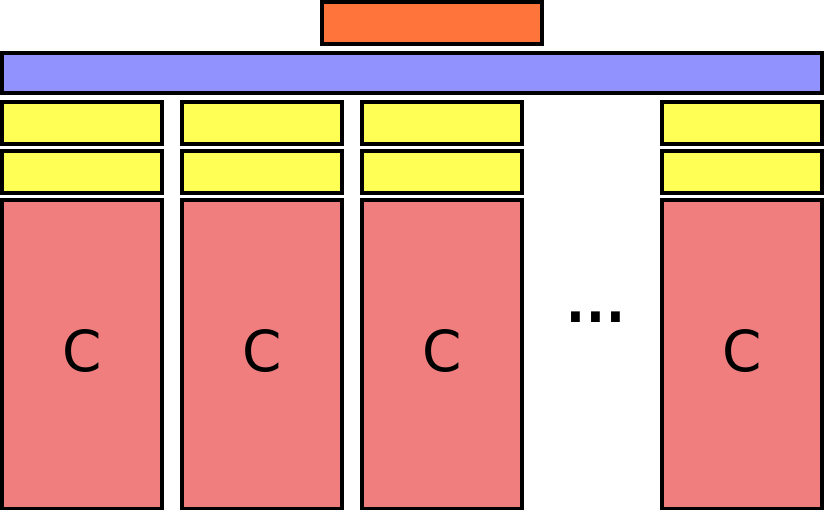}
        \caption{Late Pooling}
    \end{subfigure}
    \begin{subfigure}[ht]{0.48\linewidth}
        \includegraphics[width=0.97\linewidth]{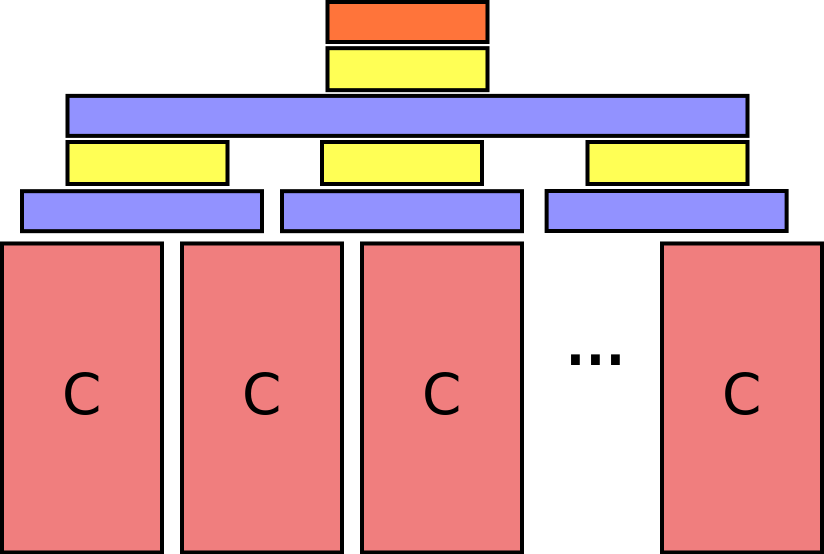}
        \caption{Slow Pooling}
    \end{subfigure}
    \begin{subfigure}[ht]{0.48\linewidth}
        \vspace{6px}
        \includegraphics[width=0.97\linewidth]{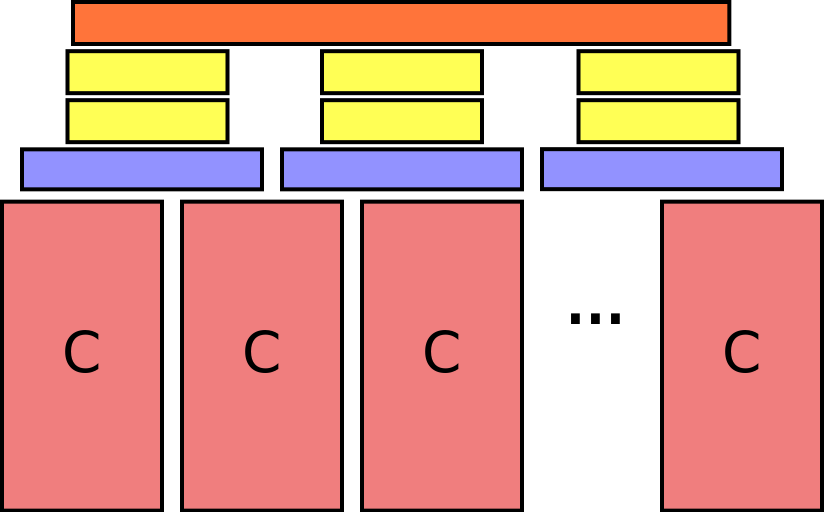}
        \caption{Local Pooling}
    \end{subfigure}
    \begin{subfigure}[ht]{0.48\linewidth}
        \vspace{6px}
        \includegraphics[width=0.97\linewidth]{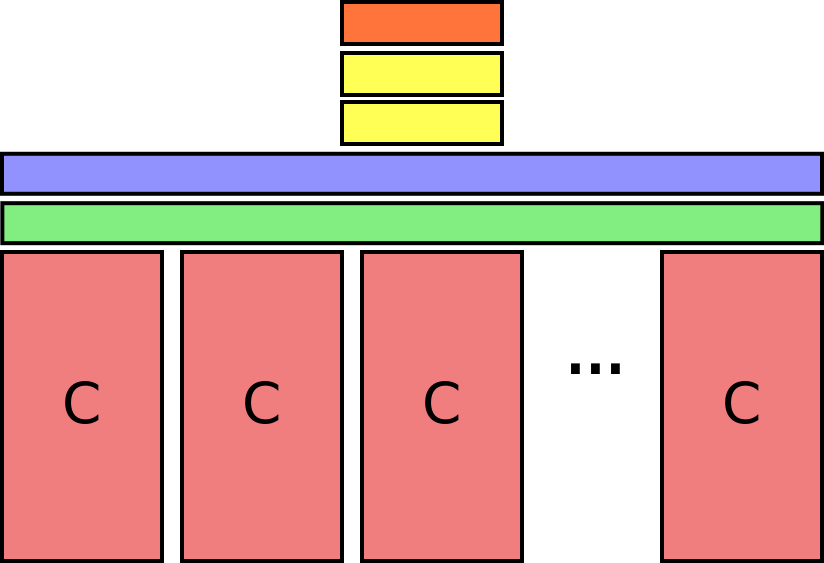}
        \caption{Time-Domain Convolution}
    \end{subfigure}

\end{center}
\caption{Different Feature-Pooling Architectures: The stacked convolutional layers are denoted by ``C''.
Blue, green, yellow and orange rectangles represent max-pooling, time-domain convolutional, fully-connected and softmax layers respectively.}
\label{fig:maxpool}
\end{figure}

{\bf Conv Pooling:}
The Conv Pooling model performs max-pooling over the final convolutional layer across the video's frames. A key advantage of this network is
that the spatial information in the output of the convolutional layer is preserved through a max operation over the time domain.

{\bf Late Pooling:} The Late Pooling model first passes convolutional features through two fully connected layers before applying the max-pooling
 layer. The weights of all convolutional layers and fully connected layers are shared.
 Compared to Conv Pooling, Late Pooling directly combines high-level information across frames.

{\bf Slow Pooling:} Slow Pooling hierarchically combines frame level information from smaller temporal windows. Slow Pooling uses a two-stage pooling strategy: max-pooling is first applied over 10-frames of convolutional features with stride 5 (e.g. max-pooling may be thought of as a size-10 filter being convolved over a 1-D input with stride 5). Each max-pooling layer is then followed by a fully-connected layer with shared weights. In the second stage, a single max-pooling layer combines the outputs of all fully-connected layers. In this manner, the Slow Pooling network groups temporally local features before combining high level information from many frames.

{\bf Local Pooling:} Similar to Slow Pooling, the Local Pooling model combines frame level features locally after the last convolutional layer. Unlike Slow Pooling, Local Pooling only contains a single stage of max-pooling after the convolutional layers. This is followed by two fully connected layers, with shared parameters. Finally a larger softmax layer is connected to all towers. By eliminating the second max-pooling layer, the Local Pooling network avoids a potential loss of temporal information.

{\bf Time-Domain Convolution:} The Time-Domain Convolution model contains an extra time-domain convolutional layer before feature pooling across frames. Max-pooling is performed on the temporal domain after the time-domain convolutional layer. The convolutional layer consist of 256 kernels of size $3 \times 3$ across 10 frames with frame stride 5. This model aims at capturing local relationships between frames within a small temporal window.

{\bf GoogLeNet Conv Pooling}: We experimented with an architecture based on GoogLeNet~\cite{szegedy14going}, in which the max-pooling operation
is performed after the dimensionality reduction (average pooling) layer in GoogLeNet. This is the layer which in the original architecture was
directly connected to the softmax layer. We enhanced this architecture by adding two fully connected
layers of size 4096 with ReLU activations on top of the 1000D output but before softmax. Similar to AlexNet-based models, the weights of convolutional layers
 and inception modules are shared across time.

\vspace{-.6em}
\subsection{LSTM Architecture}
\vspace{-.6em}

In contrast to max-pooling, which produces representations which are order
invariant, we propose using a recurrent neural network to explicitly consider
\textit{sequences} of CNN activations. Since videos contain dynamic content,
the variations between frames may encode additional information which could be
useful in making more accurate predictions.

Given an input sequence $\textbf{x} = (x_1,\dotsc,x_T)$ a standard recurrent
neural network computes the hidden vector sequence $\textbf{h} =
(h_1,\dotsc,h_T)$ and output vector sequence $\textbf{y} =
(y_1,\dotsc,y_T)$ by iterating the following equations from $t = 1$ to
$T$:

\begin{align}
  h_t &= \mathcal{H}(W_{ih}x_t + W_{hh}h_{t-1} + b_h) \\
  y_t &= W_{ho}h_t + b_o
\end{align}

\noindent where the $W$ terms denote weight matrices (e.g. $W_{ih}$ is the
input-hidden weight matrix), the $b$ terms denote bias vectors
(e.g. $b_h$ is the hidden bias vector) and $\mathcal{H}$ is the hidden
layer activation function, typically the logistic sigmoid function.

\begin{figure}[t]
\begin{center}
\includegraphics[width=0.95\linewidth]{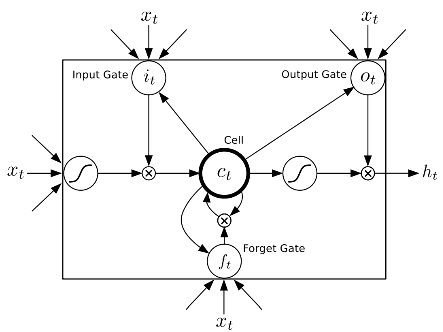}
\end{center}
\caption{Each LSTM cell remembers a single floating point value $c_t$ (Eq. \ref{eqc}). This value may be diminished or erased through a multiplicative interaction with the forget gate $f_t$ (Eq. \ref{eqf}) or additively modified by the current input $x_t$ multiplied by the activation of the input gate $i_t$ (Eq. \ref{eqi}). The output gate $o_t$ controls the emission of $h_t$, the stored memory $c_t$ transformed by the hyperbolic tangent nonlinearity (Eq. \ref{eqo},\ref{eqh}). Image duplicated with permission from Alex Graves.}
\label{fig:lstm_cell}
\end{figure}

Unlike standard RNNs, the Long Short Term Memory (LSTM) architecture
\cite{gers02learning} uses memory cells (Figure \ref{fig:lstm_cell})
to store and output information, allowing it to better discover long-range
temporal relationships. The hidden layer $\mathcal{H}$ of the
LSTM is computed as follows:

\begin{align}
i_t &= \sigma (W_{xi}x_t + W_{hi}h_{t-1} + W_{ci}c_{t-1} + b_i) \label{eqi}\\
f_t &= \sigma (W_{xf}x_t + W_{hf}h_{t-1} + W_{cf}c_{t-1} + b_f) \label{eqf}\\
c_t &= f_tc_{t-1} + i_t\;\textrm{tanh}(W_{xc}x_t + W_{hc}h_{t-1} + b_c) \label{eqc}\\
o_t &= \sigma (W_{xo}x_t + W_{ho}h_{t-1} + W_{co}c_t + b_o) \label{eqo}\\
h_t &= o_t\;\textrm{tanh}(c_t) \label{eqh}
\end{align}

\noindent where $\sigma$ is the logistic sigmoid function, and $i$, $f$, $o$,
and $c$ are respectively the \textit{input gate}, \textit{forget
  gate}, \textit{output gate}, and \textit{cell} activation
vectors. By default, the value stored in the LSTM cell $c$ is
maintained unless it is added to by the input gate $i$ or diminished
by the forget gate $f$. The output gate $o$ controls the emission of
the memory value from the LSTM cell.

\begin{figure}[ht]
\begin{center}
\includegraphics[width=0.95\linewidth]{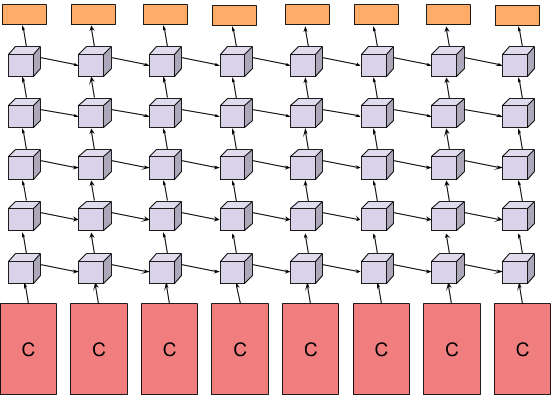}
\end{center}
\caption{Deep Video LSTM takes input the output from the final CNN
  layer at each consecutive video frame. CNN outputs are processed
  forward through time and upwards through five layers of stacked
  LSTMs. A softmax layer predicts the class at each time step.
  The parameters of the convolutional networks (pink) and softmax
  classifier (orange) are shared across time steps.}
\label{fig:lstm-arch}
\end{figure}

\label{subsec:lstm-arch}
We use a deep LSTM architecture \cite{graves13speech} (Figure
\ref{fig:lstm-arch}) in which the output from one LSTM layer is input
for the next layer. We experimented with various numbers of layers
and memory cells, and chose to use five stacked LSTM layers, each
with 512 memory cells. Following the LSTM layers, a Softmax classifier
makes a prediction at every frame.

\vspace{-.6em}
\subsection{Training and Inference}
\vspace{-.6em}
\label{subsec:training}
The max-pooling models were optimized on a cluster using Downpour Stochastic Gradient Descent
starting with a learning rate of $10^{-5}$ in conjunction with a momentum of 0.9 and weight decay of 0.0005.
For LSTM, we used the same optimization method with a learning rate of $N *
10^{-5}$ where $N$ is number of frames.  The learning rate was exponentially
decayed over time. Each model had between ten and fifty replicas split across
four partitions. To reduce CNN training time, the parameters of AlexNet and
GoogLeNet were initialized from a pre-trained ImageNet model and then
fine-tuned on Sports-1M videos.

\textbf{Network Expansion for Max-Pooling Networks:}
\label{subsec:expansion}
Multi-frame models achieve higher accuracy at the cost of longer
training times than single-frame models.
Since pooling is performed after CNN towers that
share weights, the parameters for a single-frame and multi-frame
max-pooling network are very similar.
This makes it possible to expand a single-frame model to a
multi-frame model. Max-pooling models are first
initialized as single-frame networks then expanded to 30-frames and
again to 120-frames. While the feature distribution of the max-pooling
layer could change dramatically as a result of expanding to a larger
number of frames (particularly in the single-frame to 30-frame case),
experiments show that transfering the parameters is nonetheless
beneficial. By expanding small networks into larger ones and then
fine-tuning, we achieve a significant speedup compared to training a
large network from scratch.

\textbf{LSTM Training:} We followed the same procedure as training max-pooled
network with two modifications: First,  the
video's label was backpropagated at each frame rather than once per clip.
Second, a gain $g$ was applied to the gradients backpropagated at each frame.
$g$ was linearly interpolated from $0...1$ over frames $t=0...T$. $g$ had the
desired effect of emphasizing the importance of correct prediction at later
frames in which the LSTM's internal state captured more information. Compared
empirically against setting $g=1$ over all time steps or setting $g=1$ only at
the last time step $T$ ($g=0$ elsewhere), linearly interpolating $g$ resulted
in faster learning and higher accuracy. For the final results, during training
the gradients are backpropagated through the convolutional layers for fine
tuning.

\textbf{LSTM Inference:} In order to combine LSTM frame-level predictions into a single video-level
prediction, we tried several approaches: 1) returning the prediction at
the last time step $T$, 2) max-pooling the predictions over time,
3) summing the predictions over time and return the max 4)
linearly weighting the predictions over time by $g$ then sum and return
the max.

The accuracy for all four approaches was less than $1\%$ different,
but weighted predictions usually resulted in the best performance,
supporting the idea that the LSTM's hidden state becomes progressively
more informed as a function of the number of frames it has seen.

\vspace{-.6em}
\subsection{Optical Flow}
\vspace{-.6em}
Optical flow is a crucial component of any video classification approach because it encodes the pattern of
apparent motion of objects in a visual scene.
 Since our networks process video frames at $1fps$, they do not use any apparent motion information.
Therefore, we additionally train both our temporal models on optical flow
 images and perform late fusion akin to the two-stream hypothesis proposed by~\cite{simonyan2014two}.

 Interestingly, we found that initializing from a model trained on raw image
 frames can help classify optical flow images by allowing faster convergence
 than when training from scratch. This is likely due to the fact that features that can describe for raw
 frames like edges also help in classifying optical flow images. This is
 related to the effectiveness of Motion Boundary Histogram (MBH), which is analogous to computing Histogram of Oriented Gradients (HOG) on
 optical flow images, in action recognition~\cite{wang13}.

Optical flow is computed from two adjacent frames sampled at $15 fps$ using the approach of~\cite{zach07}.
To utilize existing implementation and networks trained on raw frames, we store optical flow as images
by thresholding at $-40, 40$ and rescaling the horizontal and vertical components of the flow to $[0,255]$ range.
The third dimension is set to zero when feeding to the network so that it gives no effect on learning and inference.

 In our investigation, we treat optical flow in the same fashion as image frames to learn global description of videos using both feature pooling and LSTM
networks.

\vspace{-1em}
\section{Results}
\vspace{-1em}
\label{sec:results}
We empirically evaluate the proposed architectures on the Sports-1M and UCF-101 datasets with the goals of
investigating the performance of the proposed architectures, quantifying the effect of the number of frames and frame rates on classification
performance, and understanding the importance of motion information through optical flow models.

\vspace{-.6em}
\subsection{Sports-1M dataset}

\vspace{-.6em}
\label{subsec:sports-1m}
The Sports-1M dataset \cite{karpathy2014large} consists of roughly 1.2 million
YouTube sports videos annotated with 487 classes, and it is representative of
videos in the wild. There are 1000-3000 videos per class and approximately
$5\%$ of the videos are annotated with more than one class. Unfortunately,
since the creation of the dataset, about 7\% of the videos have been removed by
users. We use the remaining 1.1 million videos for the experiments below.

Although Sports-1M is the largest publicly available video dataset,
the annotations that it provides are at video level. No information is
given about the location of the class of interest. Moreover, the videos
in this dataset are unconstrained. This means that the camera movements
are not guaranteed to be well-behaved, which means that unlike UCF-101,
where camera motion is constrained, the optical flow quality varies
wildly between videos.

\textbf{Data Extraction:} The first 5 minutes of each video are sampled
at a frame rate of $1 fps$ to obtain 300 frames per video.
Frames are repeated from the start for videos that are shorter than 5 minutes.
We learn feature pooling models that process up to 120 frames (2 minutes of
video) in a single example.

\textbf{Data Augmentation:} Multiple examples per video are obtained by
randomly selecting the position of the first frame and consistent random crops of each
frame during both training and testing. It is necessary to ensure that the same transforms
are applied to all frames for a given start/end point. We process all images in the
chosen interval by first resizing them to $256 \times 256$ pixels, then randomly
sampling a $220 \times 220$ region and randomly flipping the image horizontally
with $50\%$ probability. To obtain predictions for a video we randomly sample
$240$ examples as described above and average all predictions, unless noted otherwise.
Since LSTM models trained on a fixed number of frames can generalize to any number of frames,
we also report results of using LSTMs without data augmentation.

\textbf{Video-Level Prediction:} Given the nature of the methods presented in this paper,
it is possible to make predictions for the entire video without needing to sample, or aggregate (
the networks are designed to work on an unbounded number of frames for prediction). However,
for obtaining the highest possible classification rates, we observed that it is best to only
do this if resource constrained (i.e., when it is only possible to do a single pass over the
video for prediction). Otherwise the data augmentation method proposed above yields between 3-5\%
improvements in Hit@1 on the Sports-1M dataset.

\textbf{Evaluation:} Following \cite{karpathy2014large}, we use Hit@k values, which indicate the fraction of test
 samples that contain at least one of the ground truth labels in the top k predictions. We provide both video level
and clip level Hit@k values in order to compare with previous results where clip hit is the hit on a single video clip
 (30-120 frames) and video hit is obtained by averaging over multiple clips.

\textbf{Comparison of Feature-Pooling Architectures:}
Table~\ref{table:pooling} shows the results obtained using the
different feature pooling architectures on the Sports-1M dataset when
using a 120 frame AlexNet model. We find that max-pooling over the
outputs of the last convolutional layer provides the best clip-level
and video-level hit rates. Late Pooling, which max-pools after the
fully connected layers, performs worse than all other methods,
indicating that preserving the spatial information while performing
the pooling operation across the time domain is important.
Time-Domain Convolution gives inferior results compared to max-pooling models.
This suggests that a single time-domain convolutional layer is not effective in learning
temporal relations on high level features, which motivates us to explore more sophisticated
network architectures like LSTM which learns from temporal sequences.

\begin{center}
  \begin{table}[t]
\hfill{}
\scalebox{0.95}{
\begin{tabular}{|l|c|c|c|}
\hline
Method & Clip Hit@1 & Hit@1 & Hit@5\\
\hline\hline
Conv Pooling & \textbf{68.7} & \textbf{71.1} & \textbf{89.3} \\
Late Pooling & 65.1 & 67.5 & 87.2 \\
Slow Pooling & 67.1 & 69.7 & 88.4 \\
Local Pooling & 68.1 & 70.4 & 88.9 \\
\hline
Time-Domain  & \multirow{2}{*}{64.2} & \multirow{2}{*}{67.2} & \multirow{2}{*}{87.2} \\
 Convolution &  &  &  \\
\hline
\end{tabular}
}
\hfill{}
\vspace{-1em}
\caption{Conv-Pooling outperforms all other feature-pooling architectures (Figure \ref{fig:maxpool}) on Sports-1M using a 120-frame AlexNet model.}
\label{table:pooling}
\end{table}
\end{center}

\vspace{-1em}
\textbf{Comparison of CNN Architectures:}
\label{subsec:cnnbaselines}
AlexNet and GoogLeNet single-frame CNNs (Section \ref{subsec:cnns})
were trained from scratch on single-frames selected at random from
Sports-1M videos. Results (Table \ref{table:cnn-pairings}) show that
both CNNs outperform Karpathy \etal's prior single-frame models
\cite{karpathy2014large} by a margin of 4.3-5.6\%. The increased
accuracy is likely due to advances in CNN architectures and sampling
more frames per video when training (300 instead of 50).

Comparing AlexNet to the more recent GoogLeNet yields a
1.9\% increase in Hit@5 for the max-pooling architecture, and
an increase of 4.8\% for the LSTM. This is roughly comparable to a
 4.5\% decrease in top-5 error moving from the Krizhevsky-style
CNNs that won ILSVRC-13 to GoogLeNet in ILSVRC-14. For the max-pool architecture,
this smaller gap between architectures is likely caused by the increased number
of noisy images in Sports-1M compared to ImageNet.

\begin{table}[t]
\begin{center}
  \scalebox{0.95}{
\begin{tabular}{|l|c|c|}
\hline
Method & Hit@1 & Hit@5 \\ %
\hline\hline
AlexNet single frame & 63.6 & 84.7 \\ %
GoogLeNet single frame & \textbf{64.9} & \textbf{86.6} \\ %
\hline
LSTM + AlexNet (fc) & 62.7 & 83.6 \\ %
LSTM + GoogLeNet (fc) & \textbf{67.5} & \textbf{87.1} \\ %
\hline
Conv pooling + AlexNet & 70.4 & 89.0 \\ %
Conv pooling + GoogLeNet & \textbf{71.7} & \textbf{90.4} \\ %
\hline
\end{tabular}
}
\end{center}
\vspace{-1em}
\caption{
  GoogLeNet outperforms AlexNet alone and when paired with both Conv-Pooling
  and LSTM. Experiments performed on Sports-1M using 30-frame Conv-Pooling and
  LSTM models. Note that the (fc) models updated only the final layers while training and did not use data augmentation.}
\label{table:cnn-pairings}
\end{table}

\begin{table}[t]
\begin{center}
\scalebox{0.95}{
\begin{tabular}{|l|c|c|c|c|}
\hline
Method & Frames & Clip Hit@1 & Hit@1 & Hit@5\\
\hline\hline
LSTM & 30 & N/A & 72.1 & 90.4 \\
\hline
\multirow{2}{*}{Conv pooling} & 30 & 66.0 & 71.7 & 90.4 \\
& 120 & 70.8 & 72.3 & 90.8 \\
\hline
\end{tabular}
}
\end{center}
\vspace{-1em}
\caption{Effect of the number of frames in the model.
Both LSTM and Conv-Pooling models use GoogLeNet CNN.}
\label{table:frames}
\end{table}

\begin{table}
\begin{center}
\scalebox{0.95}{
\begin{tabular}{|l|c|c|}
\hline
Method & Hit@1 & Hit@5\\
\hline\hline
LSTM on Optical Flow & 59.7 & 81.4 \\
LSTM on Raw Frames & 72.1 & 90.6 \\
LSTM on Raw Frames + & \multirow{2}{*}{73.1} & \multirow{2}{*}{90.5} \\
LSTM on Optical Flow &  & \\
\hline
30 frame Optical Flow & 44.5 & 70.4 \\
Conv Pooling on Raw Frames & 71.7 & 90.4 \\
Conv Pooling on Raw Frames + & \multirow{2}{*}{71.8} & \multirow{2}{*}{90.4} \\
Conv Pooling on Optical Flow &  & \\
\hline
\end{tabular}
}
\end{center}
\vspace{-1em}
\caption{Optical flow is noisy on Sports-1M and if used alone, results in lower
performance than equivalent image-models. However, if used in conjunction with
raw image features, optical flow benefits LSTM. Experiments performed on
30-frame models using GoogLeNet CNNs.} \label{table:sports-optflow}
\end{table}

\begin{table*}
\begin{center}
\begin{tabular}{|l|c|c|c|c|c|}
\hline
Category & Method & Frames & Clip Hit@1 & Hit@1 & Hit@5\\
\hline\hline
Prior & Single Frame & 1 & 41.1 & 59.3 & 77.7 \\
Results \cite{karpathy2014large} & Slow Fusion & 15 & 41.9 & 60.9 & 80.2 \\
\hline
Conv Pooling & Image and Optical Flow & 120 & {\bf 70.8} & 72.4 & {\bf 90.8} \\
\hline
LSTM & Image and Optical Flow & 30 & N/A & {\bf 73.1} & 90.5 \\
\hline
\end{tabular}
\vspace{-1em}
\caption{Leveraging global video-level descriptors, LSTM and Conv-Pooling
  achieve a 20\% increase in Hit@1 compared to prior work on the in Sports-1M
dataset. Hit@1, and Hit@5 are computed at video level.} \label{table:sports_results}
\end{center}
\vspace{-2em}
\end{table*}

\textbf{Fine Tuning:} When initializing from a pre-trained network, it is not
always clear whether fine-tuning should be performed. In our experiments, fine
tuning was crucial in achieving high performance. For example, in
Table~\ref{table:cnn-pairings} we show that a LSTM network paired with
GoogLeNet, running on 30 frames of the video achieves a Hit@1 rate of 67.5. However,
the same network with fine tuning achieves 69.5 Hit@1. Note that these results do not use
data augmentation and classify the entire 300 seconds of a video.

\textbf{Effect of Number of Frames:}
Table~\ref{table:frames} compares Conv-Pooling and LSTM models as a function of the number of frames aggregated. In terms of clip hit, the 120 frame model
 performs significantly better than the 30 frame model.
Also our best clip hit of $70.8$ represents a $70\%$ improvement over the Slow Fusion approach of \cite{karpathy2014large} which uses
clips of few seconds length. This confirms our initial hypothesis that we need to consider the entire video in order to benefit
more thoroughly from its content.

\textbf{Optical Flow:}
Table~\ref{table:sports-optflow} shows the results of fusion with the optical
flow model. The optical flow model on its own has a much lower accuracy
(59.7\%) than the image-based model (72.1\%) which is to be expected given that
the Sports dataset consists of YouTube videos which are usually of lower
quality and more natural than hand-crafted datasets such as UCF-101. In the
case of Conv Pooling networks the fusion with optical flow has no significant
improvement in the accuracy.  However, for LSTMs the optical flow model is able
to improve the overall accuracy to $73.1\%$.

\textbf{Overall Performance:}
Finally, we compare the results of our best models against the previous
state-of-art on the Sports-1M dataset at the time of submission.
Table~\ref{table:sports_results} reports the results of the best model
from~\cite{karpathy2014large} which performs several layers of 3D convolutions
on short video clips against ours.  The max-pool method shows an increase of
18.7\% in video Hit@1, whereas the LSTM approach yields a relative increase of 20\%.
The difference between the max-pool and LSTM method is explained by the fact
that the LSTM model can use optical flow in a manner which lends itself to late
model fusion, which was not possible for the max-pool model.

\vspace{-.6em}
\subsection{UCF-101 Dataset}
\vspace{-.6em}
\label{subsec:ucf-101} The UCF-101~\cite{ucf101} contains 13,320 videos
with 101 action classes covering a broad set of activities such as sports,
musical instruments, and human-object interaction.  We follow the suggested
evaluation protocol and report the average accuracy over the given three
training and testing partitions. It is difficult to train a deep network with
such a small amount of data. Therefore, we test how well our models that are
trained in Sports-1M dataset perform in UCF-101.

\textbf{Comparison of Frame Rates:} Since UCF-101 contains short
videos, 10-15 seconds on average, it is possible to extract frames at
higher frame rates such as $6 fps$ while still capturing context from
the full video. We compare 30-frame models trained at three different
frame-rates: $30 fps$ (1 second of video) and $6 fps$ (5 seconds). Table
\ref{table:frame_rate} shows that lowering the frame rate from $30 fps$ to $6
fps$ yields slightly better performance since the model obtains more context
from longer input clips.  We observed no further improvements when decreasing
the frame rate to $1 fps$. Thus, as long as the network sees enough context
from each video, the effects of lower frames rate are marginal. The LSTM model,
on the other hand can take full advantage of the fact that the videos can be
processed at 30 frames per second.

\textbf{Overall Performance:} Our models achieve state-of-the-art
performance on UCF-101 (Table \ref{table:optical_flow_results}),
slightly outperforming approaches that use hand-crafted features and
CNN-based approaches that use optical flow. As before, the performance
edge of our method results from using increased numbers of frames
to capture more of the video.

Our 120 frames model improves upon previous work~\cite{simonyan2014two} ($82.6\%$ vs $73.0\%$) when considering models that learn directly from raw frames
 without optical flow information. This is a direct result of considering larger context within a video, even when the frames within a short clip are
 highly similar to each other.

Compared to Sports-1M, optical flow in UCF-101
provides a much larger improvement in accuracy ($82.6\%$ vs. $88.2\%$ for max-pool).
This results from UCF-101 videos being better centered, less shaky, and better
trimmed to the action in question than the average YouTube video.

\textbf{High Quality Data:} The UCF-101 dataset contains short,
well-segmented videos of concepts that can typically be identified in a single
frame. This is evidenced by the high performance of single-frame networks
(See Table \ref{table:optical_flow_results}). In contrast, videos in the wild often
feature spurious frames containing text or shot transitions, hand-held video
shot in either first person or third person, and non-topical segments such as
commentators talking about a game.

\begin{table}[t]
\begin{center}
\scalebox{0.85}{
\begin{tabular}{|c|c|c|}
\hline
Method & Frame Rate & 3-fold Accuracy (\%) \\
\hline
\hline
Single Frame Model & N/A & 73.3 \\
\hline
\multirow{2}{*}{Conv Pooling (30 frames)} & 30~fps & 80.8 \\
                                          & 6~fps & 82.0 \\
\hline
\multirow{2}{*}{Conv Pooling (120 frames)} & 30~fps & 82.6 \\
                                           & 6~fps & 82.6 \\
\hline
\end{tabular}
}
\end{center}
\vspace{-1em}
\caption{Lower frame rates produce higher UCF-101 accuracy for 30-frame Conv-Pooling models.}
\label{table:frame_rate}
\vspace{-2em}
\end{table}

\begin{table}[t]
\begin{center}
\scalebox{0.85}{
\begin{tabular}{|p{7cm}|p{2cm}|}
\hline
Method & 3-fold Accuracy (\%) \\
\hline\hline
Improved Dense Trajectories (IDTF)s~\cite{wang13} & 87.9 \\
\hline
Slow Fusion CNN \cite{karpathy2014large} & 65.4 \\
\hline
Single Frame CNN Model (Images) \cite{simonyan2014two} & 73.0 \\
\hline
Single Frame CNN Model (Optical Flow) \cite{simonyan2014two} & 73.9 \\
\hline
Two-Stream CNN (Optical Flow + Image Frames, Averaging) \cite{simonyan2014two} & 86.9 \\
\hline
Two-Stream CNN (Optical Flow + Image Frames, SVM Fusion) \cite{simonyan2014two} & 88.0 \\
\hline
\hline
Our Single Frame Model & 73.3 \\
\hline
Conv Pooling of Image Frames + Optical Flow (30 Frames) & 87.6 \\
Conv Pooling of Image Frames + Optical Flow (120 Frames) & \textbf{88.2} \\
\hline
\hline
LSTM with 30 Frame Unroll (Optical Flow + Image Frames) & {\bf 88.6} \\
\hline
\end{tabular}
}
\end{center}
\vspace{-1em}
\caption{UCF-101 results. The bold-face numbers represent results that are higher than previously reported results.}
\label{table:optical_flow_results}
\vspace{-2em}
\end{table}

\section{Conclusion}
\vspace{-1em}
We presented two video-classification methods capable of
aggregating frame-level CNN outputs into video-level predictions:
Feature Pooling methods which max-pool local information through time
and LSTM whose hidden state evolves with each subsequent frame. Both
methods are motivated by the idea that incorporating information across
longer video sequences will enable better video classification. Unlike
previous work which trained on seconds of video, our networks utilize
up to two minutes of video (120 frames) for optimal classification performance.
If speed is of concern, our methods can process an entire video in one shot.
Training is possible by expanding smaller networks into progressively larger
ones and fine-tuning. The resulting networks
achieve state-of-the-art performance on both the Sports-1M and UCF-101
benchmarks, supporting the idea that learning should take place over the entire
video rather than short clips.

Additionally, we explore the necessity of motion information, and
confirm that for the UCF-101 benchmark, in order to obtain
state-of-the-art results, it is necessary to use optical
flow. However, we also show that using optical flow is not always
helpful, especially if the videos are taken from the wild as is the
case in the Sports-1M dataset. In order to take advantage of optical
flow in this case, it is necessary to employ a more sophisticated
sequence processing architecture such as LSTM. Moreover, using LSTMs
on both image frames, and optical flow yields the highest published
performance measure for the Sports-1M benchmark.

In the current models, backpropagation of gradients proceeds down
all layers and backwards through time in the top layers, but
not backwards through time in the lower (CNN) layers. In the future,
it would be interesting to consider a deeper integration of the temporal sequence information
 into the CNNs themselves. For instance, a Recurrent Convolutional Neural Network may
be able to generate better features by utilizing its own activations
in the last frame in conjunction with the image from the current
frame.

{\small
\bibliographystyle{ieee}
\bibliography{egbib}
}

\end{document}